# FSP-Diff: Full-Spectrum Prior-Enhanced Dual-Domain Latent Diffusion for Ultra-Low-Dose Spectral CT Reconstruction

Peng Peng[+], Xinrui Zhang[+], Junlin Wang, Lei Li, Shaoyu Wang*,Qiegen Liu, *Senior Member, IEEE*

*Abstract*—**Spectral computed tomography (CT) with photon-counting detectors holds immense potential for material discrimination and tissue characterization. However, under ultra-low-dose conditions, the sharply degraded signal-to-noise ratio (SNR) in energy-specific projections poses a significant challenge, leading to severe artifacts and loss of structural details in reconstructed images. To address this, we propose FSP-Diff, a full-spectrum prior–enhanced dual-domain latent diffusion framework for ultra-low-dose spectral CT reconstruction. Our framework integrates three core strategies: 1) Complementary Feature Construction: We integrate direct image reconstructions with projection-domain denoised results. While the former preserves latent textural nuances amidst heavy noise, the latter provides a stable structural scaffold to balance detail fidelity and noise suppression. 2) Full-Spectrum Prior Integration: By fusing multi-energy projections into a high-SNR full-spectrum image, we establish a unified structural reference that guides the reconstruction across all energy bins. 3) Efficient Latent Diffusion Synthesis: To alleviate the high computational burden of high-dimensional spectral data, multi-path features are embedded into a compact latent space. This allows the diffusion process to facilitate interactive feature fusion in a lower-dimensional manifold, achieving accelerated reconstruction while maintaining fine-grained detail restoration. Extensive experiments on simulated and real-world datasets demonstrate that FSP-Diff significantly outperforms state-of-the-art methods in both image quality and computational efficiency, underscoring its potential for clinically viable ultra-low-dose spectral CT imaging.**

## I. INTRODUCTION

SPECTRAL computed tomography (CT) based on photon-counting detectors (PCDs) has shown great potential in lesion detection, tissue characterization, and material decomposition, owing to its capability to accurately resolve X-ray attenuation information across multiple energy levels [1, 2]. However, as the spectral range is subdivided into narrower energy bins, the photon flux per bin decreases substantially. This leads to a precipitous decline in the signal-to-noise ratio (SNR) of energy-specific projections, thereby severely degrading the quality of reconstructed images. Furthermore, given the inherent health risks associated with ionizing radiation [3], the total exposure cannot be arbitrarily increased to compensate for these losses. Consequently, under clinically acceptable dose constraints, energy-specific reconstructions often suffer from pronounced artifacts and noise. Therefore, achieving high-quality and high-fidelity spectral CT reconstruction under low-dose protocols has emerged as a critical challenge in medical imaging research.

Traditional CT reconstruction predominantly relies on analytical algorithms like filtered backprojection (FBP) [4]. Although FBP excels in computational efficiency, its performance degrades sharply under ultra-low-dose conditions due to its sensitivity to noise. To overcome this limitation, iterative reconstruction algorithms incorporate image priors (e.g., dictionary learning [5] and total variation (TV) [6]) to suppress noise and preserve details. In spectral CT, advanced models like PRISM [7], joint-channel total nuclear variation [8], and nonlocal low-rank and sparse matrix decomposition [9]. have been developed to exploit cross-channel correlations. Despite their theoretical rigor, these methods are often hampered by intensive computational overhead and a reliance on manually tuned parameters, which limit their practicality in time-sensitive clinical settings [10].

In recent years, deep learning has revolutionized CT image reconstruction. Supervised methods, notably FBPConvNet, which refines initial FBP images via a multi-resolution U-Net [11], and RED-CNN, which employs a residual encoder–decoder architecture with shortcut connections [12] establish end-to-end mapping to significantly enhance low-dose image quality. Researchers have since expanded CNN-based approaches through sophisticated architectures, incorporating residual blocks [13-15], generative adversarial networks (GANs) [16, 17], attention mechanisms [18-20], and knowledge distillation [21], among others [22, 23]. Despite their success, these models rely heavily on large-scale paired datasets and often generalize poorly across different scan protocols. Furthermore, their "black-box" nature lacks explicit modeling of CT physics, potentially limiting their clinical interpretability.

As emerging unsupervised generative models, diffusion models learn underlying data distributions and have demon-

This work was supported in part bythe National Natural Science Foundation of China under Grant U24A20304, Grant U25A20407 and Grant 62201616. (Peng Peng and Xinrui Zhang are co-first authors.) (Corresponding author:Shaoyu Wang.)

P. Peng, S. Wang, and Q. Liu are with School of Information Engineering, Nanchang University, Nanchang 330031, China (e-mail: 6105123139@email.ncu.edu.cn; {wangshaoyu, liuqiegen}@ncu.edu.cn).

J. Wang is with School of Mathematics and Computer Sciences, Nanchang University, Nanchang 330031, China. (e-mail: 7812123133@email.ncu.edu.cn).

X. Zhang, and L. li are with the Henan Key Laboratory of Imaging and Intelligent Processing, Information Engineering University, Zhengzhou 450001, China (e-mail: zxr200043@163.com; leehotline@163.com).

strated significant potential in image generation and restoration tasks [24, 25]. Principled frameworks like Denoising Diffusion Probabilistic Models (DDPM) [26] and SDE [27] offer robust capabilities in modeling complex distributions and solving inverse problems. To mitigate the high computational costs inherent in their iterative nature, several acceleration strategies have been developed. DDIM [28] and CoreDiff [29] reduce inference steps, while RDDM [30] and latent diffusion model (LDM) [31] simplify representation through residual or latent-domain modeling. Furthermore, inverse-problem-driven approaches such as IRSDE [32] guide the reverse diffusion with physical or data-consistency constraints to accelerate convergence. Despite these advances, applying standard diffusion models to ultra-low-dose spectral CT remains challenging. The high-dimensional nature of spectral data entails significant redundancy, characterized by spatial sparsity within individual energy bins and strong structural correlations across energy channels. However, existing methods predominantly focus on distribution modeling within a single domain, neglecting these intrinsic spectro-physical attributes. Furthermore, they often overlook the heterogeneous information distributions across different imaging domains and fail to fully exploit structural and prior information available from other domains, thereby limiting reconstruction performance.

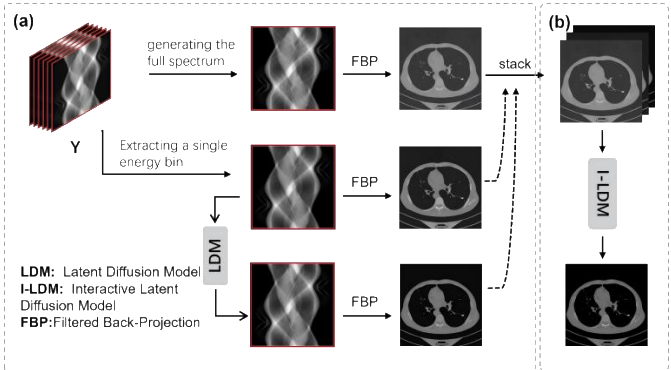

Fig. 1. The simplified overview of the FSP-Diff framework. (a) Multipath information generation, (b) Multipath Interactive Translation.

To address these technical bottlenecks, we fully exploit and integrate the complementary information from multiple sources. As shown in Fig. 1, the FSP-Diff framework establishes a unified reconstruction paradigm that effectively leverages heterogeneous information sources through joint interactive reconstruction while maintaining high computational efficiency. The principal contributions of this work are summarized as follows:

1) Full-spectrum-guided dual-domain latent diffusion: We propose a latent diffusion framework that explicitly leverages inter-spectral correlations and integrates complementary priors from both projection and image domains. By performing diffusion within a unified, low-dimensional latent space, the model significantly enhances reconstruction efficiency and practical feasibility for high-dimensional ultra-low-dose spectral CT.

2) Dual-domain collaborative feature construction: We exploit domain-dependent information distributions to construct synergistic representations. Specifically, projection-domain denoising provides stable structural scaffolds and global topological consistency, while direct image-domain reconstruction preserves rich textural nuances. This collaborative approach ensures an optimal balance between high-SNR structural integrity and fine-grained detail fidelity.

3) Full-spectrum prior integration for structural enhancement: To fully exploit the prior information within spectral data, we fuse low-SNR energy-specific projections to construct a high-SNR full-spectrum prior image. This composite image serves as a unified structural reference to guide the reconstruction of individual energy bins, effectively enhancing textural fidelity and overall image quality under extreme noise conditions.

## II. RELATED WORK

### A. Latent Space Encoding

Encoder-decoder models form a cornerstone of deep learning, extensively applied to data compression, feature extraction, and generative modeling. The autoencoder (AE) [33] established a foundational framework wherein an encoder compresses input data into a compact latent representation, and a decoder reconstructs the data from it. While effective for dimensionality reduction, AEs often learn simplistic approximations, limiting their ability to capture complex data patterns.

To enhance robustness, the denoising autoencoder (DAE) [34] was introduced. A DAE deliberately corrupts the input with noise and is trained to recover the original, clean data. This forces the model to learn more meaningful and invariant features, improving performance on noisy or partial inputs.

Subsequently, the variational autoencoder (VAE) [35], incorporated a probabilistic framework, learning a distribution over the latent space rather than a fixed representation. This advancement enables VAEs to not only perform feature extraction but also generate novel data samples.

A significant architectural shift arrived with the Transformer model [36], which utilizes self-attention mechanisms to capture long-range dependencies in data. It has since become a general-purpose backbone for diverse sequence modeling tasks.

BuDenoising Diffusion Probabilistic Modelsilding on these concepts, LDM [31] enhances generative modeling efficiency by conducting the diffusion process within a lower-dimensional latent space. This approach leverages an autoencoder for latent representation, balancing high computational efficiency with superior output quality in image generation.

### B. Denoising Diffusion Probabilistic Models

Diffusion Models constitute a class of likelihood-based generative models that learn to transform a simple distribution (e.g., Gaussian noise) into a complex data distribution through a gradual denoising process. The framework consists of two stages: a fixed forward process that progressively corrupts data with noise, and a learned reverse process that recovers the data.

The forward process is defined as a fixed Markov chain that systematically introduces noise to the data over a series of $T$ timesteps. The transition between successive states in this chain is governed by the following equation:

$$q(x_t \mid x_{t-1}) = \mathcal{N}(x_t; \sqrt{1-\beta_t}x_{t-1}, \beta_t I), \qquad (1)$$

where $x_t$ denotes the intermediate state image at timestep $t$, $\beta_t$ is a predefined variance schedule controlling the noise scale at step $t$, and $\mathcal{N}$ represents the Gaussian distribution.

In the reverse process, the model initializes the generation with a Gaussian random noise map $x_T$ and applies a learned denoising function to progressively transform it into a high-quality output $x_0$, formally expressed as:

$$p(x_{t-1} \mid x_t, x_0) = \mathcal{N}(x_{t-1}; \mu_t(x_t, x_0), \sigma_t^2 I), \quad (2)$$

where the mean $\mu_t(x_t, x_0) = \frac{1}{\sqrt{\alpha_t}}\left(x_t - \epsilon \frac{1-\alpha_t}{\sqrt{1-\bar{\alpha}_t}}\right)$ and the variance $\sigma_t^2 = \frac{1 \cdot \bar{\alpha}_{t-1}}{1 \cdot \bar{\alpha}_t}\beta_t$. $\epsilon$ represents the noise present in $x_t$, which is the critical variable to be estimated. A denoising network $\epsilon_\theta(x_t, t)$ is trained to predict this noise.

## III. METHOD

### A. Motivation

In recent years, diffusion models have shown considerable promise for modeling and image reconstruction in spectral CT. Existing approaches have sought to better exploit inherent priors in spectral data—for example, by leveraging inter-energy correlations for multi-channel joint modeling and posterior sampling [37], applying masked projection-domain architectures that enable high- and low-energy interaction [38], or deploying alternating iterative diffusion frameworks across multiple domains [39, 40]. However, under ultra-low-dose conditions, these approaches exhibit notable limitations. They typically fail to concurrently address the modeling of high-dimensional data characteristics, incorporate spectral correlation priors, and leverage complementary information from multiple domains. The severely degraded signal-to-noise ratio fundamentally impedes the effective extraction of usable information, consequently making the reconstruction of high-quality, spectrally consistent images exceptionally challenging. Furthermore, the direct implementation of the diffusion process in pixel space incurs substantial computational costs when handling multi-energy data, which significantly restricts its practical applicability.

High-dimensional spectral CT data inherently contain richer redundant information compared to conventional CT data. In the single-energy domain, such data often exhibit sparsity and low-rank properties, while in the multi-energy domain, they demonstrate strong structural correlations. Moreover, data distributions in the projection and image domains differ significantly, leading to distinct representations of information. Therefore, a critical challenge in low-dose spectral CT reconstruction lies in the effective integration of these multiple classes of priors to achieve high-quality image reconstruction. Furthermore, building on our previous work [41], we discovered that high-quality full-spectrum images can be synthesized from single-energy inputs. These full-spectrum images encompass broad spectral information and possess a higher signal-to-noise ratio, offering a strong structural prior that can serve as guidance to enhance ultra-low-dose imaging quality.

Motivated by these considerations, we develop a full-spectrum prior-enhanced dual-domain latent diffusion frame-work for ultra-low-dose spectral CT reconstruction. As shown in Fig .2, owing to domain-dependent information distributions, features with complementary information can be generated through dual-domain collaborative processing. Subsequently, by fusing low-SNR, energy-specific projections, we construct a high-SNR full-spectrum prior image that guides the reconstruction process across all energy bins. Moreover, to alleviate the high computational burden caused by high-dimensional spectral data and multi-class prior information, the multi-path features are embedded into a compact latent space, thereby enabling the diffusion process to perform interactive feature fusion on a low-dimensional manifold.

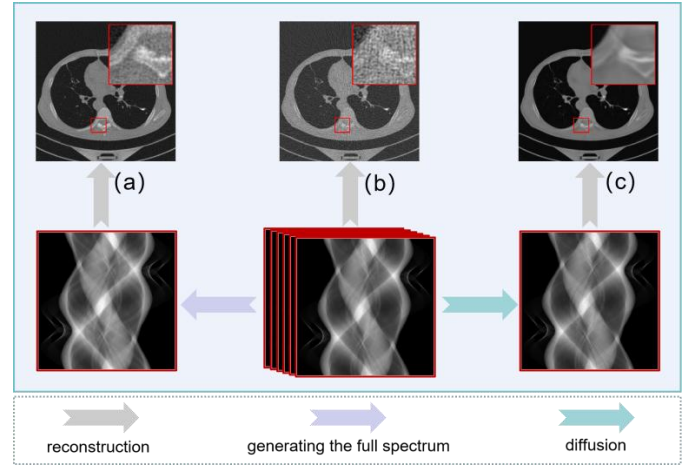

Fig. 2. Reconstructions derived from different projection types. (a) The full-spectrum reconstructed image, (b) The energy-bin reconstructed image, (c) The denoised energy-bin reconstructed image.

### B. Training Process of Our Diffusion Method

Although the proposed framework contains two domain-specific diffusion instantiations, both follow the same two-stage training paradigm consisting of encoder and decoder pretraining and latent space diffusion training.

In the first stage (Fig. 3), a latent feature space projector is trained to compress input images into latent representations that preserve essential features, thereby accelerating computation. In the second stage, the diffusion model is trained to learn the prior distribution of the latent representations encoded by the pretrained projector from the first stage.

**Encoder and Decoder Pretraining:** Following prior work [42], the latent feature space projector is instantiated as an encoder–decoder pair. The encoder—primarily composed of stacked residual blocks and linear layers—compresses the input into a highly compact latent representation; concretely, a $512 \times 512$ image is mapped to a vector of length $4C$, where $C$ denotes the number of learned features. The Image Reconstruction (IR) Transformer Decoder (IRTD) integrates a dynamic Transformer block with a U-shaped architecture to reconstruct the image from the latent representations. The dynamic Transformer comprises Modulated Channel-Transposed Attention (MCTA) and Modulated Gated Feed-Forward Networks (MGFN), with the latent representation serving as dynamic modulation parameters to preserve recovered details. As illustrated in Fig. 3, the encoder and decoder are optimized jointly during pretraining. Before encoding, the ground-truth (GT) pixel-space data $P_{GT}$ and the low-quality (LQ) pixel-space data $P_{LQ}$ are multiplied by a weight matrix $w$ to modulate the overall value range of the image space, thereby facilitating feature extraction by the encoder.

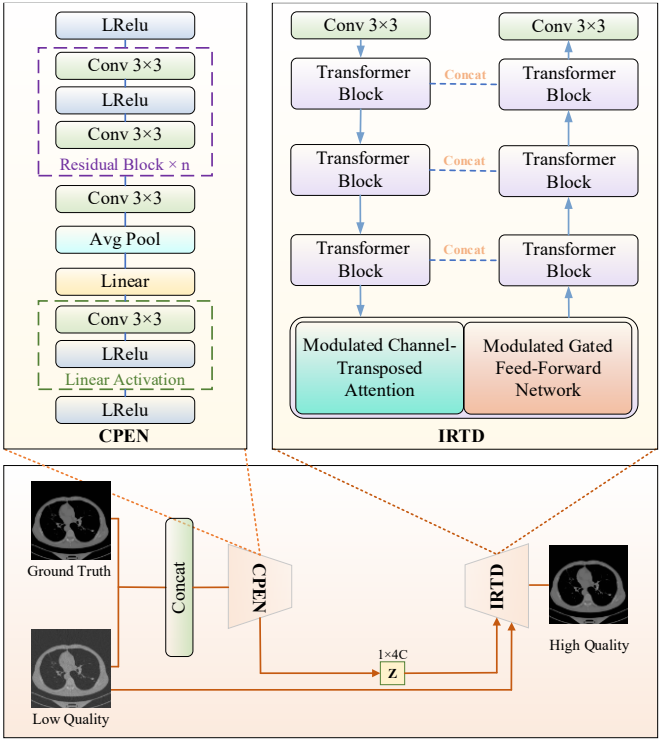

Fig. 3. The training of CIPE and IRTD. The CIPE extracts global feature information from the image, while the IRTD generates the corresponding image by decoding information from the global features.

After weighting, the $P_{GT}$ and $P_{LQ}$ pixel-space images are concatenated and downsampled via a PixelUnshuffle operator to form the Compact IR Prior Encoder (CIPE) input. The CIPE then maps this input to a compact latent representation $Z$. For clarity, we write

$$Z = E\left(U\left(Concat\left(P_{GT}, P_{LQ}\right)\right)\right),\qquad(3)$$

where $E(\cdot)$ is the encoder and $U(\cdot)$ denotes PixelUnshuffle. $D(\cdot)$ which denotes IRTD is trained to map the latent representation back to pixel-space, yielding a high-quality (HQ) image $P_{HQ}$.

$$P_{HQ} = D\left(Z, P_{LQ}\right).\qquad(4)$$

Because the CIPE and IRTD are optimized jointly with the objective of recovering the original image from a lower-dimensional representation, the encoder learns a projection that incurs minimal information loss. The overall pretraining loss, denoted as $\mathcal{L}_{res}$, is formulated as a composite function comprising an $\mathcal{L}_1$ loss and an $\mathcal{L}_{ssim}$ loss:

$$\mathcal{L}_{res} = \parallel P_{GT} - P_{HQ} \parallel_1 + \lambda \parallel P_{GT} - P_{HQ} \parallel_{ssim}.\qquad(5)$$

**Latent-space Diffusion Training:** The second stage trains a diffusion model to estimate the IR prior (IPR). The weighted $P_{GT}$ and $P_{LQ}$ images are first passed through the pretrained CIPE to obtain the IPR; we denote this target by $Z_0$. A forward DDPM process then perturbs $Z_0$ according to a variance

schedule over $T$ steps, gradually transforming the data distribution into Gaussian noise:

$$q(Z_t \mid Z_{t-1}) = \mathcal{N}(Z_t; \sqrt{1-\beta_t}Z_{t-1}, \beta_t I).\qquad(6)$$

Eq. (6) can be further simplified as follows:

$$q(Z_T \mid Z_0) = \mathcal{N}\left(Z_T; \sqrt{\overline{\alpha_T}}Z_0, (1-\overline{\alpha_T})I\right),\qquad(7)$$

where $\mathcal{N}(\cdot)$ denotes the Gaussian distribution. $\alpha_T$ and $\overline{\alpha}_T$ are defined as: $\alpha_T = 1 - \beta_T$, $\overline{\alpha}_T = \prod_{i=0}^{T} \alpha_i$, $\beta_{1:T} \in (0,1)$ control the injected noise level at step $t$. During diffusion training the denoiser is conditioned on the measured latent extracted from the $P_{LQ}$ by CIPE:

$$B = E\left(U\left(P_{LQ}\right)\right).\qquad(8)$$

The diffusion model is optimized to predict and remove the injected noise on noisy latent $Z_t$. while beging conditioned on the measured latent $B$. The training objective function for the diffusion models is written as

$$\mathcal{L}_{diff} = \mathbb{E}_{(x), B, \eta \sim \mathcal{N}(0,I), t}[\parallel \eta - \eta_\theta(Z_t, t, B) \parallel_1],\qquad(9)$$

where $Z_t$ is the noisy latent representation at step $t$, $B$ denotes the measured latent representation, and $\eta_\theta$ is the parameterized diffusion network. During inference, a Gaussian latent $Z_T \sim \mathcal{N}(0, I)$ is sampled and progressively denoised by subtracting the predicted noise $\eta$ at each step to obtain a clean latent representation $Z_0$ which is subsequently decoded by IRTD to generate the high-quality pixel-space image.

The multi-path inputs are jointly encoded into high-dimensional spatial feature representations via a unified encoder. Rather than processing channels independently, all inputs are embedded collectively, yielding an expanded feature tensor with increased channel dimensionality while retaining full spatial resolution. This joint mapping extracts shared structural patterns and complementary inter-channel information at the feature level.

These features are then progressively refined through successive Transformer blocks, forming a hierarchical set of multi-level spatial feature maps. At each level, self-attention captures global contextual dependencies, implicitly enabling inter-channel interaction within the high-dimensional feature space.

Throughout this process, spatial correspondence is strictly preserved, and features remain as dense spatial maps. To further condition these representations, an Iterative Prior Refinement (IPR) module is incorporated as a conditional modulation signal. Injected across multiple Transformer layers, the IPR dynamically recalibrates channel-wise feature responses, providing a structured feature basis for subsequent diffusion-based refinement and enabling adaptive integration of complementary cross-channel information.

### C. Reconstruction Process of Our Method

**Projection-Domain Latent Diffusion:** The first stage operates in the projection domain. Specifically, the noisy projection $y^n$ is encoded by CIPE1 into a latent representation $B^p$, which subsequently serves as the conditional input to the latent diffusion model.

Based on the inversion of the forward process in DDPM as defined in (2), the complete projection image space data can be recovered from the learned primary structure distribution. The sampling procedure is described as follows: beginning with Gaussian noise $Z_T^P \sim \mathcal{N}(0, I)$, the diffusion model iteratively refines the latent variable $Z_T^P$ using the learned function $\eta_\theta^P(Z_t^P, t, B^P)$, where $B^P$ denotes the conditional vector. This process continues until $t = 1$, producing the final output $Z_0^P$. The overall sampling process can be formally expressed as follows:

$$Z_{T-1}^P = \frac{1}{\sqrt{\alpha_t}}\left(Z_t^P - \frac{1 - \alpha_t}{\sqrt{1 - \overline{\alpha}_t}}\eta_\theta(Z_t^P, t, B^P)\right). \quad (10)$$

The compact latent representation allows for accurate estimation with fewer iterations and a smaller model than traditional diffusion approaches. However, without joint training of the denoising process and the IRTD1, minor estimation errors can prevent the IRTD1 from achieving its optimal performance. Thus, following diffusion updates, the output is decoded by the IRTD1 to recover projection-domain data. The following objective function is used for joint training of both phases:

$$\mathcal{L}_{all}^P = \mathcal{L}_{res}^P + \mathcal{L}_{diff}^P. \quad (11)$$

The decoding process yields denoised projection $y_0^n$, which is subsequently reconstructed into image $x_T$ via FBP.

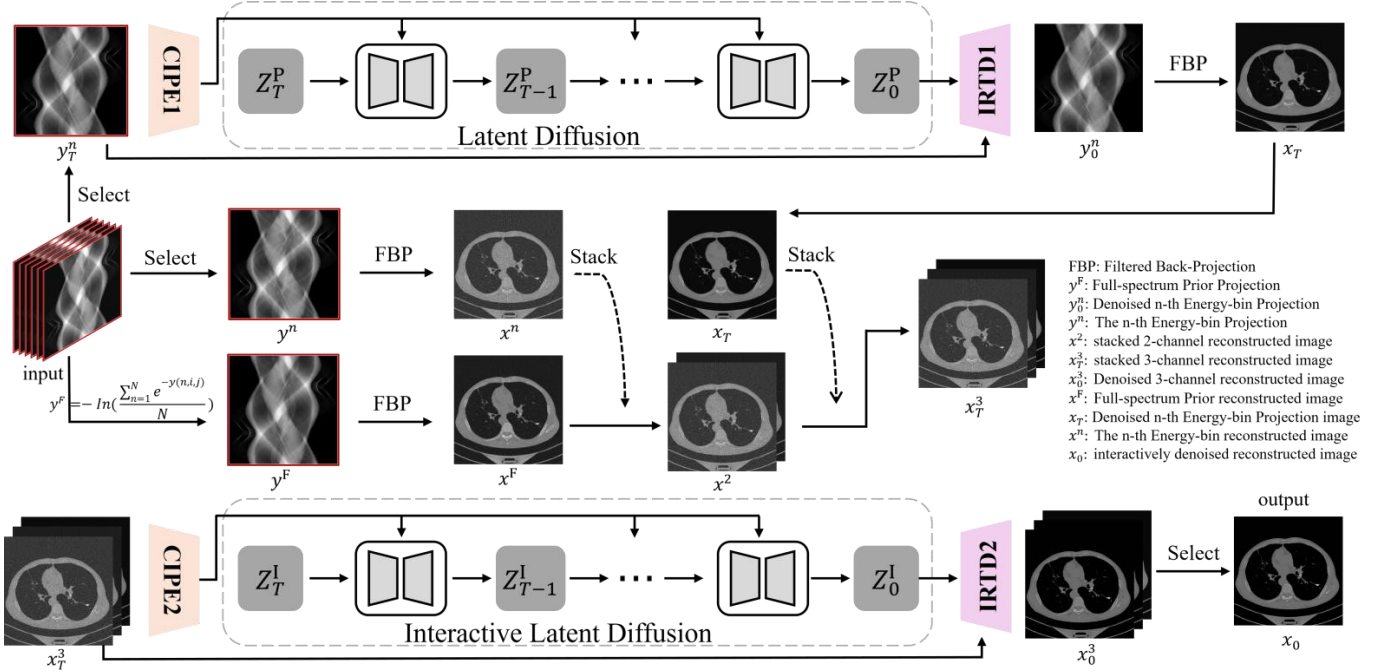

Fig. 4. The pipeline of the reconstruction procedure in FSP-Diff. The noisy projection is first denoised using a latent diffusion model and subsequently reconstructed (top). Subsequently, reconstruction is independently performed from the noisy projections and the fused full-spectrum projections generated via noisy projection fusion (middle). The resulting three reconstructed images are finally integrated and refined via an interactive latent diffusion model to produce the final reconstruction (bottom).

**Full-Spectrum Prior Construction:** To enhance reconstruction quality, a high-SNR full-spectrum prior projection image is generated by integrating information across all energy bins using a proposed energy-fusion algorithm. This full-spectrum reconstructed image serves as a guiding prior, providing reliable structural and textural information to the subsequent image-domain reconstruction. The following objective function is used to generate full-spectrum projection:

$$y_F = -In(\frac{\sum_{n=1}^N e^{-y(n,i,j)}}{N}), \quad (12)$$

where $y^F$ denotes the synthesized full-spectrum projection. N is the total number of energy bins in the spectral CT acquisition. The coordinates $(i, j)$ specify the pixel location in the projection image. The fused full-spectrum projection $y^F$ is reconstructed into the image $x^F$ via FBP.

**Image-Domain Interactive Latent Diffusion:** In this stage, the interactive latent diffusion model is fed a stacked three-channel input $x_T^3$, which comprises the single-energy FBP reconstruction $x^n$, the full-spectrum reconstructed image $x^F$, and the projection-domain denoised reconstruction $x_T$. This input is encoded by CIPE2 into a latent representation $B^I$, which subsequently guides the interactive diffusion process within a unified low-dimensional latent space.

Similar to the projection-domain reconstruction model, the entire image space data can be obtained from the learned primary structure distribution by inverting the forward process of DDPM according to (2). Sampling from Gaussian noise $Z_T^I \sim \mathcal{N}(0, I)$, and using the intermediately perturbed latent representation $Z_T^I$ and the conditional vector $B^I$ as input, the diffusion model iteratively samples $Z_{T-1}^I$ based on the learned $\eta_\theta^I(Z_t^I, t, B^I)$ until $t = 1$, yielding the final generated result $Z_0^I$. During inference, this sampling process can be described as:

$$Z_{T-1}^I = \frac{1}{\sqrt{\alpha_t}}\left(Z_t^I - \frac{1 - \alpha_t}{\sqrt{1 - \overline{\alpha}_t}}\eta_\theta(Z_t^I, t, B^I)\right). \quad (13)$$

During training, the diffusion module and the IRTD2 are jointly optimized using the loss function:

$$\mathcal{L}^1_{all} = \mathcal{L}^1_{res} + \mathcal{L}^1_{diff}. \tag{14}$$

*D. Algorithms*

---

**Algorithm 1.** FSP-Diff for reconstruction .

---

**Require:** Trained CIPE1, IRTD1 and $\eta^p_\theta(Z^p_t, t, B^p)$ . Trained CIPE2, IRTD2 and $\eta_\theta(Z^1_t, t, B^1)$ . Noisy energy spectrum projection $y^n$.

**1: Initialization:** $Z^p_T \sim \mathcal{N}(0, I), Z^1_T \sim \mathcal{N}(0, I)$

**2: Full-Spectrum Prior Construction:**

$y^F = -In(\dfrac{\sum^N_{n=1} e^{-y(n,i,j)}}{N})$

$x^F = FBP(y^F)$

**3: The Image-Domain Reconstruction:** $x^n = FBP(y^n)$

**4: Projection-Domain Diffusion:**

Encoding step: $B^p = CIPE1(y^n)$

**For** $t = T$ to 0 do

   Update $Z^p_{t-1}$ via (10)

**End for**

Decoding step: $y^n_0 \Leftarrow IRTD1(Z^p_0, x^3_T)$

$x_T = FBP(y^n_0)$

**5: Stack:** $x^F + x^n + x_T = x^3_T$

**6: Image-Domain Multi-Path Diffusion:**

Encoding step: $B^1 = CIPE2(x^3_T)$

**For** $t = T$ to 0 do

   Update $Z^1_{t-1}$ via Eq. (13)

**End for**

Decoding step: $x^3_0 \Leftarrow IRTD2(Z^1_0, x^3_T)$

**7: Return:** $x_0$

---

## IV. EXPERIMENTS

### A. Experiment Setup

To comprehensively evaluate the performance of the proposed FSP-Diff method, we carried out experiments on both simulated data and real-data. Six reconstruction methods are selected for comparison, including the conventional FBP, sparsity-enforcing TV, deep learning-enhanced FBPConvNet, and advanced diffusion-based methods CoreDiff, Denoising with Diffusion Priors (Dn-Dp) [43], and Efficient diffusion model for image restoration (DiffIR) [42]. The FSP-Diff adopts a 4-level encoder – decoder. from level 1 to level 4, the numbers of transformer blocks are [6, 5, 5, 4], and the numbers of attention heads are [1, 2, 4, 8]. For the diffusion model, the channel dimension is 64, and the total number of iteration steps is $T = 4$. To quantitatively assess the proximity of the reconstructed images to the ground truth, we adopted two widely used metrics: the Peak Signal-to-Noise Ratio (PSNR) and the Structural Similarity Index Measure (SSIM). Higher values for both PSNR and SSIM indicate superior reconstruction quality.

**Simulation Dataset:** For the simulation dataset, we generated multi-energy images for each CT slice using a discrete Beer-Lambert forward model, based on clinically verified bone and soft tissue images. This model incorporated X-ray spectra from SpekCalc [44] and mass-attenuation coefficients from the National Institute of Standards and Technology (NIST). The acquired projections were categorized into six energy-resolved channels: [52, 64], [64, 73], [73, 80], [80, 87], [87, 99], and [99, 120] keV. These CT slices were obtained from the thoracic, abdominal, and pelvic regions of the upper body using a Siemens SOMATOM Definition spectral CT scanner, located at the Radiology Department of the 988th People's Hospital. Each slice had a resolution of 512 × 512 pixels. The dataset consisted of a total of 1600 slices, with 1400 allocated for training and 200 for testing.

To simulate ultra-low-dose conditions, noisy projection data were generated by introducing Poisson noise into the clean multi-energy projections. The incident photon flux per X-ray path was set to $3 \times 10^3$ and $1.2 \times 10^4$ photons. The Siddon ray-driven algorithm [45] was employed to simulate the X-ray paths. The source-to-detector distance was set to 50 cm, with a detector width of 72 cm and a total of 512 detector elements. All projection views were evenly distributed across a full 360-degree range.

**Real Dataset:** To evaluate the generalization capability of the proposed method, we utilized a real mouse thoracic spectral CT dataset. The scan was performed on a MARS multi-energy CT system under institutional animal welfare regulations, with ethical approval from the Ethics Committee of the Li Ka Shing Faculty of Medicine at The University of Hong Kong. The acquired projections were divided into five energy-resolved channels, corresponding to [7, 32], [32, 43], [43, 54], [54, 70], and [70, 120] keV. The imaging setup involved a 156-mm source-to-object distance (SOD), a 256-mm source-to-detector distance (SDD), and a detector pixel size of 0.110 mm.

### B. Experimental Comparison

In this section, the experimental results, including both qualitative and quantitative evaluations, of different methods on the aforementioned datasets are presented.

**Simulated experiments:** In this study, multiple image slices with diverse anatomical structures and texture characteristics are randomly selected from a human CT dataset. Two noise-level scenarios are constructed by setting the number of photons per X-ray path to $3 \times 10^3$ and $1.2 \times 10^4$.

Fig. 6 and Fig. 7 present the reconstructed images and corresponding residual maps for Slice 1 (with $1.2 \times 10^4$ photons) and Slice 2 (with $3 \times 10^3$ photons), respectively. As illustrated, the filtered back-projection (FBP) method generates significant artifacts, resulting in poor reconstruction quality, particularly in regions with fine anatomical structures. Although FBPConvNet improves upon FBP by reducing some artifacts, it still retains residual noise and fails to fully recover fine details. The TV and Dn-Dp methods show some improvement: TV leads to overly smooth reconstructions, blurring critical details, while Dn-Dp manages noise reduction but struggles to restore subtle anatomical structures. CoreDiff and DiffIR offer enhanced performance, with fewer artifacts and more complete structural recovery. However, both methods still miss certain fine details. In contrast, FSP-Diff consistently delivers superior results. Even under the reduced photon count, it produces reconstructions that closely resemble the ground truth, with minimal artifacts and enhanced recovery of anatomical features, as evidenced by the residual maps.

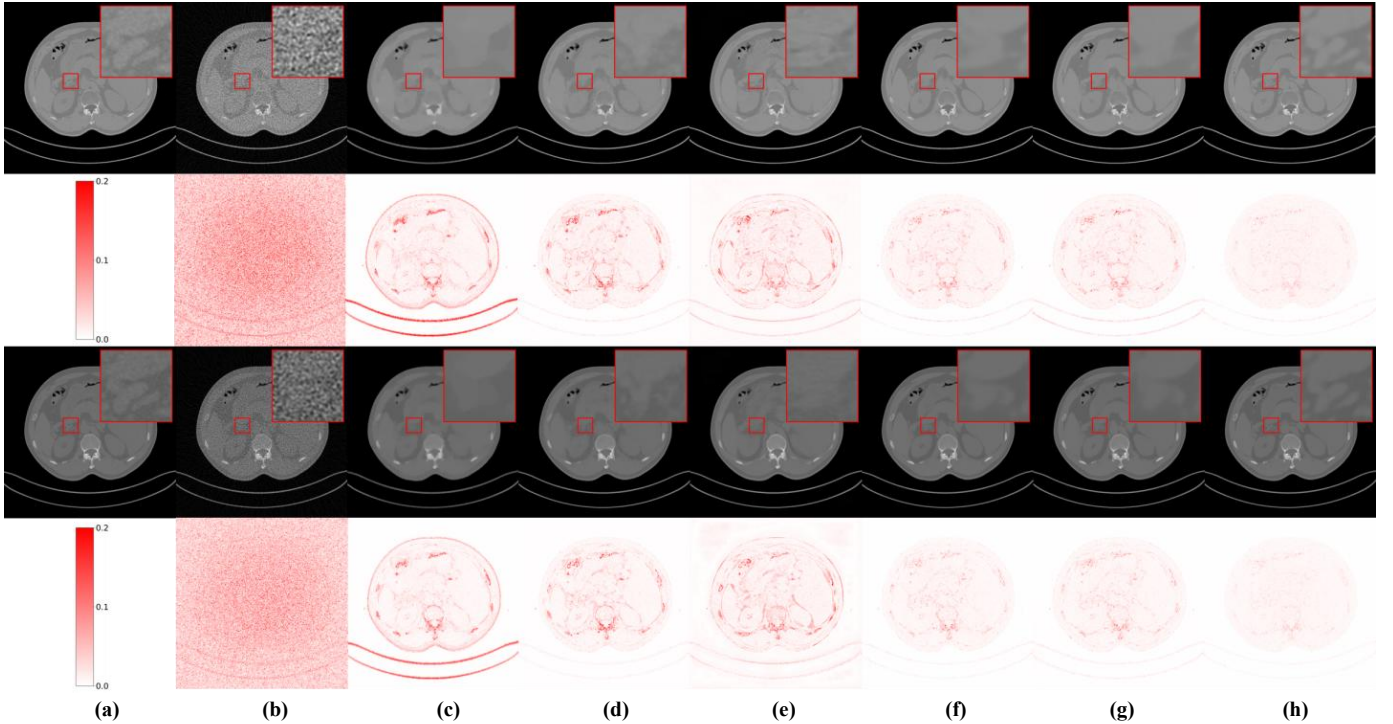

Fig. 5. Reconstruction results from the spectral dataset at $1.2 \times 10^4$ photons. The first and third rows show energy bin1 and energy bin5 reconstructed images. while the second and fourth rows show residuals between reference (window: [0,1]) and reconstructed images (residual display window: [0,0.2]). (a) The reference image, (b) FBP, (c) TV, (d) Dn-Dp, (e) FBPConvNet, (f) CoreDiff, (g) DiffIR, (h) FSP-Diff .

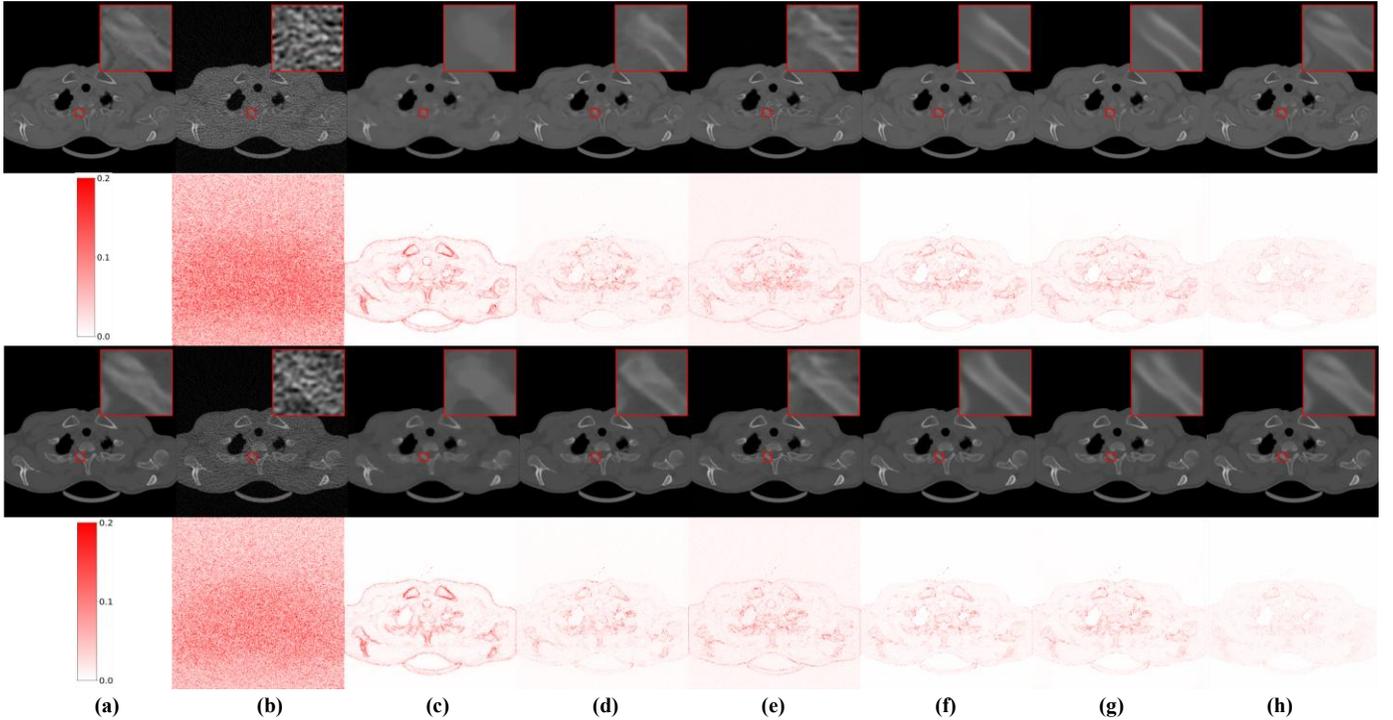

Fig. 6. Reconstruction results from the spectral dataset at $3 \times 10^3$ photons. The first and third rows show energy bin1 and energy bin5 reconstructed images. while the second and fourth rows show residuals between reference (window: [0,1]) and reconstructed images (residual display window: [0,0.2]). (a) The reference image, (b) FBP, (c) TV, (d) Dn-Dp, (e) FBPConvNet, (f) CoreDiff, (g) DiffIR, (h) FSP-Diff.

To substantiate these qualitative observations, quantitative results are provided in Tables I and Tables II, which compare the average PSNR and SSIM values of the selected slices. The best results for each photon-count setting are highlighted in bold. The data demonstrate that FSP-Diff consistently outperforms all competing methods across all spectral channels under varying photon counts. Moreover, the performance variation of FSP-Diff across energy levels is notably small, indicat-ing that the method effectively exploits the strong inter-channel correlations inherent in spectral CT acquisitions.

Together, these findings confirm the robustness, efficacy, and overall superiority of the proposed approach.

TABLE I
PSNR/SSIM Values of Spectral Images Reconstructed by Different Methods at $1.2 \times 10^4$ Photons Using the Simulation Dataset.

|  | FBP [4] | TV [6] | Dn-Dp [43] | FBPConvNet [11] | CoreDiff [29] | DiffIR [42] | **FSP-Diff** |
|---|---|---|---|---|---|---|---|
| bin1 | 28.36/0.5167 | 34.68/0.9114 | 35.68/0.9107 | 36.92/0.9507 | 37.04/0.9546 | 39.15/0.9660 | **42.10/0.9765** |
| bin2 | 29.82/0.5789 | 36.27/0.9320 | 37.28/0.9145 | 38.00/0.9578 | 39.07/0.9648 | 41.38/0.9751 | **44.61/0.9850** |
| bin3 | 29.21/0.5505 | 35.86/0.9356 | 37.61/0.9229 | 38.38/0.9616 | 39.16/0.9633 | 41.23/0.9730 | **44.69/0.9835** |
| bin4 | 29.86/0.5794 | 36.75/0.9429 | 37.32/0.9104 | 38.57/0.9637 | 40.04/0.9704 | 42.08/0.9784 | **45.06/0.9847** |
| bin5 | 30.48/0.6156 | 37.36/0.9444 | 37.25/0.9140 | 38.68/0.9654 | 39.13/0.9678 | 41.80/0.9783 | **45.35/0.9874** |
| bin6 | 29.80/0.5793 | 36.54/0.9440 | 38.27/0.9332 | 38.73/0.9668 | 40.22/0.9701 | 42.17/0.9794 | **45.62/0.9880** |

TABLE II
PSNR/SSIM Values of Spectral Images Reconstructed by Different Methods at $3 \times 10^3$ Photons Using the Simulation Dataset.

|  | FBP [4] | TV [6] | Dn-Dp [43] | FBPConvNet [11] | CoreDiff [29] | DiffIR [42] | **FSP-Diff** |
|---|---|---|---|---|---|---|---|
| bin1 | 24.99/0.3407 | 32.83/0.8988 | 33.13/0.8660 | 36.11/0.9356 | 35.85/0.9372 | 36.99/0.9545 | **39.90/0.9720** |
| bin2 | 25.99/0.3835 | 34.40/0.9214 | 35.07/0.8942 | 37.18/0.9452 | 37.86/0.9516 | 38.96/0.9660 | **42.45/0.9788** |
| bin3 | 25.44/0.3591 | 33.86/0.9235 | 34.55/0.8760 | 37.63/0.9502 | 38.11/0.9510 | 38.86/0.9639 | **42.28/0.9765** |
| bin4 | 26.05/0.3880 | 34.79/0.9326 | 35.26/0.8945 | 37.83/0.9525 | 37.92/0.9422 | 39.50/0.9702 | **43.14/0.9811** |
| bin5 | 26.68/0.4231 | 35.49/0.9340 | 35.41/0.9026 | 38.01/0.9544 | 38.73/0.9623 | 39.25/0.9693 | **42.85/0.9811** |
| bin6 | 26.00/0.3840 | 34.51/0.9346 | 35.62/0.9044 | 38.13/0.9565 | 38.99/0.9633 | 39.67/0.9716 | **43.13/0.9818** |

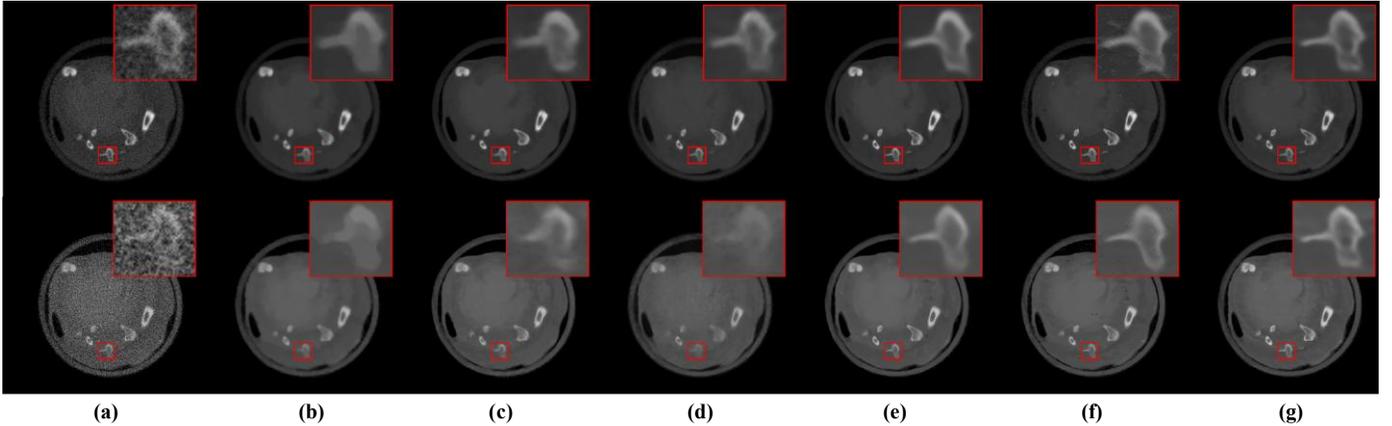

Fig. 7. Reconstruction results on the mouse thoracic spectral dataset. The first and second rows show energy bin1 and energy bin5 reconstructed images, Images reconstructed by (a) FBP, (b) TV, (c) Dn-Dp, (d) FBPConvNet, (e) DiffIR, (f) CoreDiff, (g) FSP-Diff

***Real-data experiments:*** To evaluate the generalization capability and robustness of the proposed method, knowledge learned from a human spectral CT dataset was transferred to a real-world mouse chest spectral CT dataset. Since ground-truth reconstructions are unavailable for this dataset, the evaluation was limited to qualitative comparisons. Fig. 7 presents the reconstruction results for the high-energy (top row) and low-energy (bottom row) spectral channels.

In the high-energy channel, the FBP reconstruction is dominated by severe noise, resulting in poor quality, particularly in regions with fine anatomical structures. TV regularization suppresses noise effectively, but it introduces noticeable over-smoothing and blurs structural boundaries. Both Dn-Dp and FBPConvNet further reduce noise but at the cost of blurred details and softened edges, particularly evident in the zoomed-in regions. DiffIR restores relatively clearer structures, where-as the local contrast enhancement in CoreDiff introduces visible, inhomogeneous artifacts, compromising both homogeneity and diagnostic quality. In contrast, FSP-Diff produces the sharpest reconstruction, preserving well-defined edges and fine anatomical details, demonstrating superior performance over all other methods.

The low-energy channel presents a more challenging scenario. FBP struggles to recover meaningful structures due to excessive noise. TV results in overly smoothed images, flattening textures and losing fine details. Notably, Dn-Dp performs poorly in this case, exhibiting structural degradation and inferior visual quality compared to TV. FBPConvNet suffers from a substantial loss of detail, resulting in poor accuracy. Although DiffIR and CoreDiff recover coarse anatomical outlines, residual artifacts are still visible. By contrast, FSP-Diff consistently provides clean, structurally faithful reconstruc-tions in this channel, maintaining strong structural consistency between high- and low-energy representations.

Overall, these qualitative results on the unseen mouse dataset confirm the generalization ability of FSP-Diff. The method maintains superior structural fidelity and robustness for ultra-low-dose spectral CT reconstruction across different energy levels and biological subjects.

### C. Ablation Study

To comprehensively evaluate the contribution of each component in the proposed Full-Spectrum-Enhanced Dual-Domain Collaborative Diffusion Framework (FSP-Diff), we conducted ablation experiments under the ultra–low-dose condition of $3 \times 10^3$ photons. Four model variants were examined:

1) I (Image-Domain Diffusion Only)
2) P (Projection-Domain Diffusion Only)
3) IP (Dual-Domain Collaborative Diffusion without Full-Spectrum Prior)
4) FSP-Diff (Dual-Domain Collaborative Diffusion with Full-Spectrum Prior)

All experiments were performed using the same settings on a simulated dataset with an incident photon count of $3 \times 10^3$. The quantitative results are presented visually in Fig. 8 and Fig. 9, which show the PSNR and SSIM comparisons across the various spectral bins for each model. while the corresponding visual reconstruction results are presented in Fig. 10.

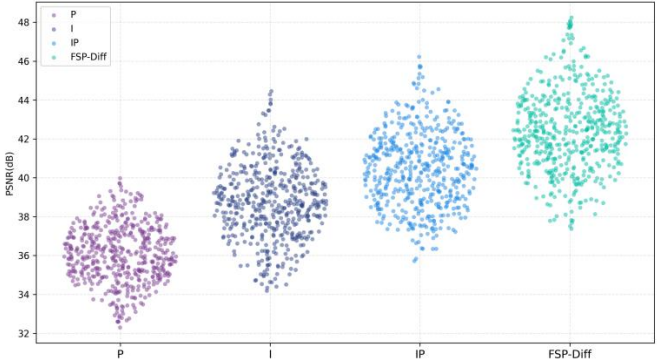

Fig. 8. PSNR comparison across different models in the ablation study.

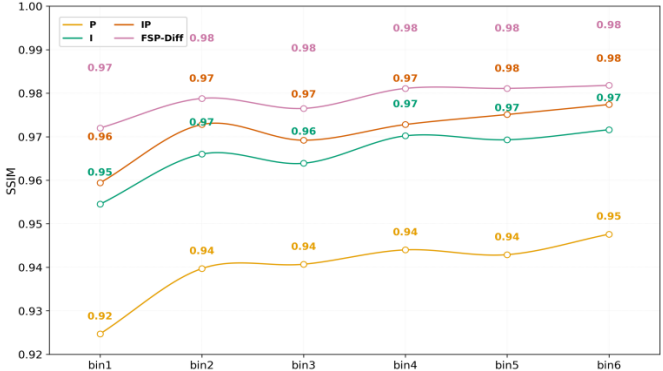

Fig. 9. SSIM comparison across different models for each energy bin in the ablation study.

Fig. 8 shows the PSNR comparison across the models. The FSP-Diff model consistently achieves the highest PSNR, highlighting its superior ability to enhance overall image quality by effectively leveraging information across multiple spectral bins. The I model, which relies solely on image-domain diffusion, outperforms the P model (projection-domain diffusion only), demonstrating a stronger ability to preserve fine structural details under severe noise. The IP model, integrating both image and projection-domain diffusion without the full-spectrum prior, shows clear improvements over the single-domain models. However, its performance still falls short of FSP-Diff, emphasizing the critical importance of the full-spectrum prior in achieving superior spectral consistency and restoring fine details.

Fig. 9 presents the SSIM comparison, which assesses the structural consistency across all six spectral bins. In this case, FSP-Diff delivers the highest SSIM, reflecting its capability to maintain structural integrity and preserve textural fidelity across the entire spectral range. The I model, while superior to P in terms of structural preservation, still falls short of FSP-Diff in maintaining cross-energy consistency. The IP model shows moderate improvement over the single-domain models but does not reach the level of structural fidelity demonstrated by FSP-Diff.

These results provide compelling evidence that both dual-domain collaboration and the full-spectrum prior are integral to achieving high-quality spectral CT reconstruction under ultra-low-dose conditions. FSP-Diff leverages these two strategies to outperform all other variants in both PSNR and SSIM, offering a substantial advantage in image quality and structural consistency.

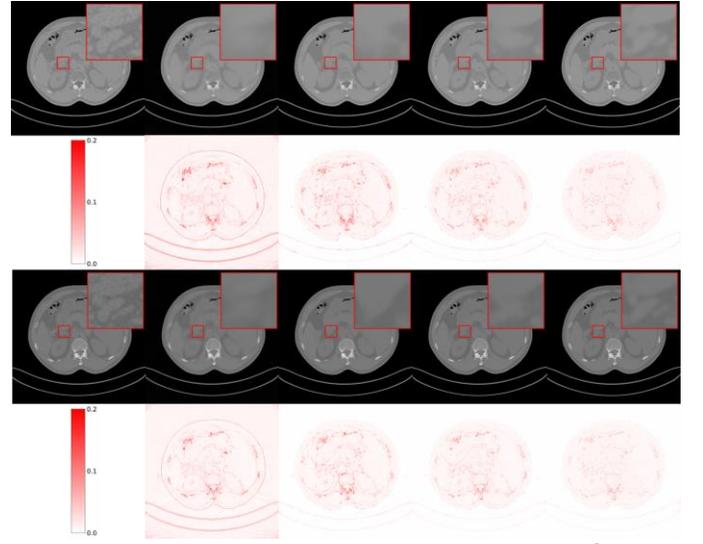

Fig. 10. Reconstruction results from the spectral dataset at $3 \times 10^3$ photons. The first and third rows show energy bin1 and energy bin5 reconstructed images. while the second and fourth rows show residuals between reference (window: [0,1]) and reconstructed images (residual display window: [0,0.2]). (a) The reference image, (b) I, (c) P, (d) IP, and (e) FSP-Diff.

Fig. 10 illustrates the reconstruction results obtained using different methods. Panel (a) shows the ground-truth clean image. Under the high-noise condition with a noise level of 3000, both the projection-domain diffusion method (b) and the image-domain diffusion method (c) exhibit pronounced over-smoothing, accompanied by the loss of fine structural details. The collaborative diffusion method between the denoised pro-jection-domain reconstruction and the noisy energy-bin image (d) further reduces the residual error, yielding a closer approximation to the target image. However, noticeable texture information remains unrecovered, and the overall improvement is limited. In contrast, the proposed enhanced dual-domain collaborative diffusion method (e) effectively denoises the image while significantly mitigating oversmoothing and cross-channel inconsistency. It preserves structural details more faithfully and results in a substantially higher-quality reconstruction.

### D. Analysis of Convergence Behavior and Computational Complexity

To evaluate the computational efficiency of the proposed diffusion model, we tested six variants using different iteration counts $T = \{1,2,4,8,16,32\}$. While Fig. 11 illustrates the corresponding PSNR variations, the primary focus of this analysis is the relationship between iteration count and processing time.

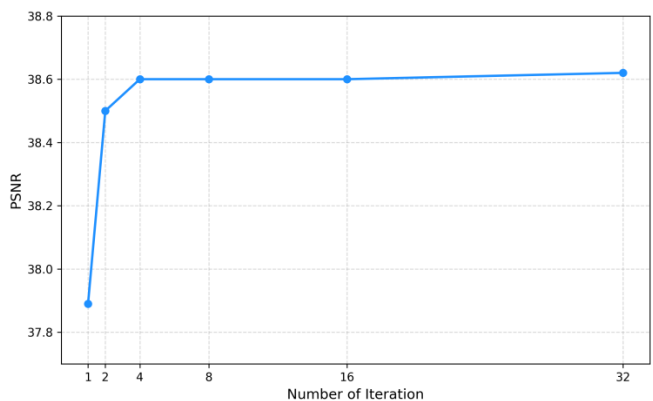

Fig. 11. PSNR variation with respect to the number of iterations.

TABLE III
Processing Time(s) per Energy Bin Comparison Among Diffusion Models

| Model | Dn-Dp | CoreDiff | DiffIR | FSP-Diff |
|-------|-------|----------|--------|----------|
| Time  | 70    | 0.17     | 0.43   | 1.22     |

Notably, when $T = 1$, the diffusion model already captures most fundamental pixel-space features and provides reasonably strong reconstruction performance. Increasing the iteration count to $T = 2$ and $T = 4$ yields further improvements, indicating that the latent feature space exhibits a compact and simple distribution, such that only a few iterations are required to extract meaningful prior information. Importantly, $T = 4$ achieves a favorable balance: it offers nearly the same reconstruction quality as larger iteration counts while maintaining significantly lower computational cost.

For higher iteration counts ($T = 8$ and $T = 16$), PSNR improvements plateau, and at $T = 32$, the performance reaches its maximum, but the gain over $T = 4$ is marginal. In contrast, the associated time cost increases substantially with larger T. Based on this efficiency–performance trade-off, we select $T = 4$ as the default iteration count, as it provides high-quality reconstructions with minimal computational burden.

A key strength of our method lies in its ability to achieve a balance between reconstruction quality and computational efficiency. As reported in Table IV, the full FSP-Diff model—despite incorporating dual domains and multi-channel spectral data—completes the reconstruction in just 1.22 seconds, demonstrating computational efficiency comparable to lightweight, single-channel diffusion models. Traditional diffusion models, on the other hand, typically require more than 200 iterations, resulting in significantly longer processing times. While some recent fast diffusion models reduce runtime, FSP-Diff not only matches these models in speed but also outperforms them in terms of image quality. The reduced iteration strategy employed in FSP-Diff accelerates inference, enabling efficient multi-channel spectral CT reconstruction without sacrificing reconstruction quality.

Compared with conventional diffusion models that typically require over 200 iterations, FSP-Diff employs a reduced-iteration strategy that significantly accelerates inference and greatly shortens reconstruction time. Although lightweight single-channel diffusion models could run slightly faster, FSP-Diff leverages dual-domain and multi-channel spectral information to more fully exploit the available measurements and reconstruct higher-quality images. As reported in Table IV, Despite incorporating dual-domain processing and multi-channel spectral CT information, FSP-Diff reconstructs high-quality images from ultra-low-dose spectral acquisitions in merely 1.22 seconds. These results demonstrate that FSP-Diff achieves an effective balance between reconstruction quality and computational efficiency.

## V. CONCLUSION

In this paper, we proposed FSP-Diff method for ultra-low-dose spectral CT reconstruction. Unlike conventional single-domain reconstruction methods, our approach integrates complementary features from both the image-domain and projection-domain. By incorporating a high-SNR full-spectrum enhancement mechanism, the framework effectively reinforces texture consistency and spectral fidelity across energy channels. Experimental results demonstrate that FSP-Diff significantly outperforms state-of-the-art methods in both objective metrics and perceptual quality, particularly under ultra-low-dose conditions. The ablation studies further validate the effectiveness of the full-spectrum enhancement and dual-domain collaboration, which together ensure robust noise suppression while preserving intricate structural and textural details. In summary, the proposed FSP-Diff framework provides a powerful and computationally efficient solution for high-quality spectral CT reconstruction under ultra-low-dose constraints, paving the way for practical applications in next-generation photon-counting CT imaging.